\documentclass[letterpaper]{article} 
\usepackage{aaai24}  
\usepackage{times}  
\usepackage{helvet}  
\usepackage{courier}  
\usepackage[hyphens]{url}  
\usepackage{graphicx} 
\usepackage{natbib}  
\usepackage{caption} 
\usepackage{algorithm}
\usepackage{algorithmic}

\usepackage{newfloat}
\usepackage{listings}
\DeclareCaptionStyle{ruled}{labelfont=normalfont,labelsep=colon,strut=off} 
\floatstyle{ruled}
\newfloat{listing}{tb}{lst}{}
\floatname{listing}{Listing}
\nocopyright

\newcommand{\primarykeywords}{
(2) Learning;
(8) Knowledge Representation/Engineering
}

\usepackage[switch, modulo]{lineno}

\title{Analyzing the Effectiveness of Large Language Models on Text-to-SQL Synthesis}
\author {
    Richard Roberson\textsuperscript{\rm 1},
    Gowtham Kaki\textsuperscript{\rm 1},
    Ashutosh Trivedi\textsuperscript{\rm 1}
}
\affiliations {
    \textsuperscript{\rm 1}University of Colorado Boulder, Boulder, Colorado, USA \\
    richard.roberson@colorado.edu, 
    gowtham.kaki@colorado.edu, 
    ashutosh.trivedi@colorado.edu
}

\begin{document}
\maketitle

\begin{abstract}
This study investigates various approaches to using Large Language Models (LLMs) for Text-to-SQL program synthesis, focusing on the outcomes and insights derived. Employing the popular Text-to-SQL dataset, \textit{spider}, the goal was to input a natural language question along with the database schema and output the correct SQL SELECT query. The initial approach was to fine-tune a local and open-source model to generate the SELECT query. After QLoRa fine-tuning WizardLM's \textit{WizardCoder}-15B model on the \textit{spider} dataset, the execution accuracy for generated queries rose to a high of 61\%. With the second approach, using the fine-tuned \textit{gpt-3.5-turbo-16k} (Few-shot) + \textit{gpt-4-turbo} (Zero-shot error correction), the execution accuracy reached a high of 82.1\%. Of all the incorrect queries, most can be categorized into a seven different categories of what went wrong: selecting the wrong columns or wrong order of columns, grouping by the wrong column, predicting the wrong values in conditionals, using different aggregates than the ground truth, extra or too few JOIN clauses, inconsistencies in the Spider dataset, and lastly completely incorrect query structure. Most if not all of the queries fall into these categories and it is insightful to understanding where the faults still lie with LLM program synthesis and where they can be improved.
\end{abstract}

\section{Introduction}

Text-to-SQL program synthesis involves converting natural language (NL) questions into SQL queries for database interaction. Historically, this task has been approached using smaller sequence-to-sequence (seq2seq) pre-trained models, which, until recently, represented the forefront of Text-to-SQL methodologies \cite{li2023resdsql}. Recent advancements in large parameter pre-trained language models, such as \textit{gpt-3.5-turbo} and gpt-4, have shifted the paradigm in Text-to-SQL research \cite{dong2023c3, gao2023texttosql}. The introduction of the \textit{spider}\footnote{https://yale-lily.github.io/spider} dataset by Yale researchers in 2019 marked a significant milestone in Text-to-SQL research \cite{yu2019spider}. These models have not only outperformed traditional seq2seq approaches but have also established new benchmarks on the \textit{spider} leaderboard.

Open-source LLMs are characterized by their publicly available model weights and the ability to undergo fine-tuning. The resulting model, named \textit{Spider Skeleton Wizard Coder}, demonstrated competitive zero-shot Text-to-SQL capabilities, rivaling those of ChatGPT (\textit{gpt-3.5-turbo}). Then, in response to challenges encountered, our research pivoted towards closed-source models. This shift was also when OpenAI introduced fine-tuning capabilities for \textit{gpt-3.5-turbo} in their API. The ability to train state-of-the-art LLMs opened new avenues for Text-to-SQL program synthesis. This paper aims to detail the strategies employed using both open and closed-source models, evaluate the results from each, and extract key insights crucial for advancing research in this field.

\section{Open Source Models}

For our first set of experiments, we chose to fine-tune WizardLM's \textit{WizardCoder}, a 15B parameter model renowned for its proficiency in programming tasks \cite{luo2023WizardCoder}. This decision was driven by the need for a coding model, especially pertinent to the \textit{spider} dataset in Text-to-SQL program synthesis. The fine-tuned model, named \textit{Spider Skeleton Wizard Coder}, attained an execution accuracy of 61\% On the \textit{spider} development set comprising 1,034 entries. The model is publicly available on the HuggingFace Transformers Library\footnote{https://huggingface.co/richardr1126/spider-skeleton-wizard-coder-merged}.

\subsubsection{Fine-tuning}
With the goal of minimizing costs, LoRA fine-tuning, denoting low-rank adapters, modifies only a fraction of the total model weights. This results in a reduced memory footprint during training. The compact LoRA is then integrated with the complete model weights, tailoring the model closer to the training data. QLoRA further minimizes the memory footprint by employing quantized weights for LoRA training, a method nearly as effective as full model fine-tuning \cite{dettmers2023qlora}.
\subsubsection{Data}
\begin{figure}[t]
\centering
\includegraphics[width=0.9\columnwidth]{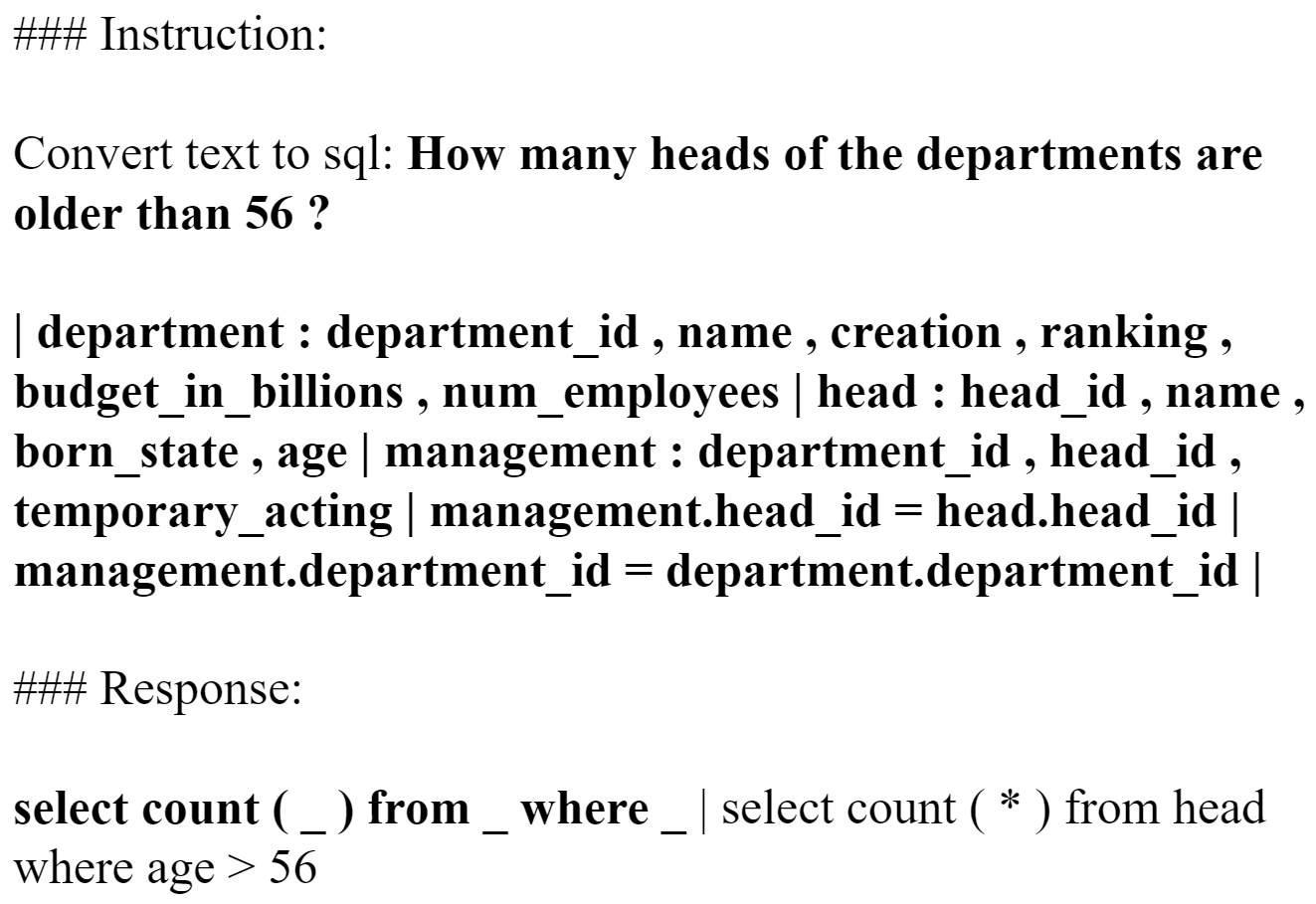} 
\caption{Shows the open-source fine-tuning dataset format for a single entry. The \textbf{bold} sections show the \textit{spider} NL question, the database schema format, and the skeleton format in the response \cite{li2023resdsql}.}
\label{fig1}
\end{figure}

The enhanced \textit{spider} dataset for fine-tuning includes the database schema for each entry, along with the skeleton format in the responses. Figure \ref{fig1} illustrates an example from this dataset. The training data response encompasses both the skeleton and the actual SQL query, prompting the model to generate responses in a similar format.

\subsubsection{Challenges}

While \textit{Spider Skeleton Wizard Coder} demonstrates commendable zero-shot execution accuracy, it still falls short. The model, fine-tuned on zero-shot data, struggles with few-shot in-context learning — the capacity to process follow-up instructions for, in this case, SQL query correction and repair \cite{brown2020fewshot}. This limitation contrasts with the capabilities of OpenAI's closed-source models like \textit{gpt-3.5-turbo}, \textit{gpt-3.5-turbo-16k}, and \textit{gpt-4-turbo}, which excel in few-shot learning due to extensive training on such data.

\section{Closed Source Models}

\begin{figure}[t]
\centering
\includegraphics[width=0.9\columnwidth]{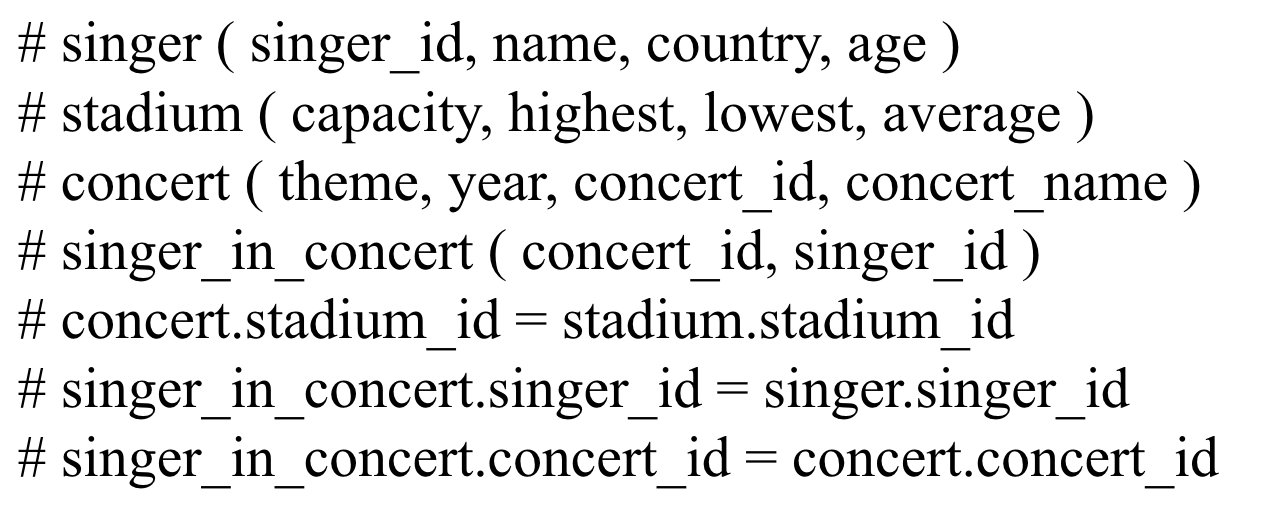} 
\caption{Shows the clear context database schema format that OpenAI suggests using for Text-to-SQL program synthesis with their models. The C3 method, zero-shot state-of-the-art, also uses this db format \cite{dong2023c3}.}
\label{fig2}
\end{figure}

For our second set of experiments, we start with the application of \textit{gpt-3.5-turbo-16k}, an extended context length variant of gpt-3.5-turbo, in a zero-shot configuration. The objective was to assess the baseline capabilities of these models prior to implementing multi-shot techniques and fine-tuning. Existing research, particularly the state-of-the-art zero-shot approach for gpt-3.5-turbo, utilizes the C3 method \cite{dong2023c3}. We follow some their methods by using a prompt alignment strategy and a database schema format. Refer to Figure \ref{fig2}.

In preliminary trials, conducted without a clear database context and prompt alignment, and conducted mainly for comparison with \textit{Spider Skeleton Wizard Coder}, the execution accuracy stood at 57.6\%. The transition from open-source to closed-source methodologies marked the adoption of more robust strategies for OpenAI's models akin to those in C3 \cite{dong2023c3}. This approach, within a zero-shot framework, achieved an execution accuracy of 68.2\%.

\subsubsection{Fine-tuning}

The subsequent phase involved fine-tuning OpenAI's \textit{gpt-3.5-turbo-16k} on the \textit{spider} dataset, incorporating database context. The fine-tuning process was facilitated using OpenAI's API. Even in a zero-shot environment, the fine-tuned model demonstrated a substantial improvement, achieving a 73.4\% execution accuracy in generating SQL queries, highlighting the effectiveness of fine-tuning models.


\subsubsection{Example Driven Correction}
The incorporation of example-driven correction, utilizing output examples to represent the expected result table after executing the queries on a real database, significantly enhanced the execution accuracy. To streamline this approach, we utilized the ground truth (gold) result table in markdown format.  Once the fine-tuned \textit{gpt-3.5-turbo-16k} generates a SQL query, its output is compared with the provided example. If there is a discrepancy, the model is informed that the output of the generated query does not align with expectations. Subsequently, the model is presented with the correct result table output example and prompted to revise the query based on this new information.

\subsubsection{Error Driven Correction}
Next, a phase known as Error Driven Correction, is activated if the SQL query, post-Example Driven Correction, still fails to align with the output example. This phase eliminates the context history from the preceding shots and establishes a fresh zero-shot environment for error correction, employing \textit{gpt-4-turbo} instead of the fine-tuned \textit{gpt-3.5-turbo-16k}. The strategy involves executing the previously generated SQL query on the actual database from the \textit{spider} dataset to detect any execution errors. Should an execution error be identified, it is incorporated into the input provided to \textit{gpt-4-turbo}. This zero-shot configuration worked better than a few-shot approach did for error correction. By integrating these correction techniques, the model's execution accuracy notably increased, reaching a peak of \textbf{82.1\%}.

\section{Insights}

The following discussion centers on the highest execution accuracy achieved: the \textit{spider} dataset fine-tuned version of \textit{gpt-3.5-turbo-16k}, supplemented by \textit{gpt-4-turbo} for error correction. This section will delve into the key insights gleaned from the SQL queries generated by this approach, particularly focusing on their comparison with the ground truth (gold) queries.

The \textit{spider} dataset can be a very challenging dataset for the extra hard queries. The 82.1\% peak execution accuracy comes from the official spider evaluation script, which groups queries based on difficulty, assigns a score for each SQL hardness category, then does a weighted average to get the final execution accuracy number. We will focus on the extra hard queries in this discussion due to this being the main area of failure in our approach.

\begin{table}[t]
\centering
\resizebox{0.9\columnwidth}{!}{
\begin{tabular}{|l|l|l|l|l|l|}
\hline
                     & \textbf{Easy} & \textbf{Medium} & \textbf{Hard} & \textbf{Extra} & \textbf{All} \\ \hline
\textbf{Count}      & 248           & 446             & 174           & 166            & 1034         \\ \hline
\textbf{Exec Accuracy} & 0.940      & 0.857           & 0.805         & 0.566          & 0.821        \\ \hline
\end{tabular}
}
\caption{Execution Accuracy by Difficulty}
\label{table1}
\end{table}

\subsection{SELECT}
Many of the incorrect queries face the issue of selecting the wrong columns in the first line of the SELECT statement. Either they use the wrong column from the selected table or it gets the column names correct but the order they are placed in doesn't match the ground truth. While the few-shot Example Driven Correction does correct a lot of these instances, they still pop up within the final predicted results.

\subsection{GROUP BY}
The \textit{spider} dataset can be challenging at times, especially in the extra hard queries. Let's take a closer look a query trying to answer the NL question: "What is the name of the course with the most students enrolled?".

\begin{listing}[h]%
\caption{Generated SQL Query}%
\label{lst:our_approach}%
\begin{lstlisting}[language=SQL]
SELECT courses.course_name
FROM student_enrolment_courses
JOIN courses ON student_enrolment_courses.course_id = courses.course_id
GROUP BY student_enrolment_courses.course_id
ORDER BY COUNT(student_enrolment_courses.student_course_id) DESC
LIMIT 1
\end{lstlisting}
\end{listing}

\begin{listing}[h]%
\caption{Ground Truth (Gold) SQL Query}%
\label{lst:ground_truth}%
\begin{lstlisting}[language=SQL]
SELECT T1.course_name
FROM Courses AS T1
JOIN Student_Enrolment_Courses AS T2 ON T1.course_id = T2.course_id
GROUP BY T1.course_name
ORDER BY count(*) DESC
LIMIT 1
\end{lstlisting}
\end{listing}

These queries are almost exactly the same, with the only difference being in the GROUP BY and ORDER BY clause. The evaluation script marks it as incorrect because both the queries return a different course with "the most students enrolled", but in reality it is just as correct as the ground truth query. The problem lies more with the \textit{spider} dataset than a problem with generation. When LIMIT 1 is removed from both of these queries the result table shows five different courses that all have the same number of students taking them. The differing GROUP BY clause is simply ordering the same results differently, so when LIMIT 1 is called, they both return different results that could both be considered correct.

\subsection{Predicting Values}
Another major problem with LLM generated SQL queries is when the LLM tries to use a conditional on a column when it doesn't know what is actually stored in that column. It is given the database schema in the format shown in Figure \ref{fig2}, but not the values stored inside those all of the table columns. For example, when trying to find the number of ships that sank in the \textit{spider} \verb|battle_death| db, the LLM might add \verb|WHERE disposition_of_ship = 'lost'| when it really should be \verb|WHERE disposition_of_ship = 'sank'|. There are no ships 'lost', they are either 'captured', 'wrecked', 'sank, or 'scuttled'. We try to correct things kinds of errors with few-shot learning and Example Driven Correction, but, in this case, the column \verb|disposition_of_ship| is not in the output result table so the Example Driven Correction shot gives no indication to the LLM as to what kinds of values are really stored in this column.

\subsection{Aggregate Columns}
The problem with using LLMs for Text-to-SQL is partially how the model interprets what the NL question is asking for. Given the NL question: "Which owner has paid for the most treatments on his or her dogs? List the owner id and last name.", our LLM approach identifies that the question is asking for which owner paid the most for their dog treatments. In the case of the ground truth query, it interprets the question as which owner brought there dog to treatment the most, and has nothing to do with the cost of treatments.

In the actual queries, the generated SQL uses a SUM aggregate in the ORDER BY clause to order the results by which owner paid the most for all pet treatments, causing the whole query to be incorrect. The ground truth SQL just uses a COUNT aggregate. And since the question only asks for id and last name to be selected, the aggregate column SUM or COUNT never gets shown in the output result table, which is why it never gets corrected in Example Driven Correction.

\subsection{JOIN}
A substantial problem arises in the generated SQL when the LLM predicts that it needs to JOIN on an extra table in order to get what the NL question is asking for. Let's take a look at this SQL answering the NL question: "What is the model for the car with a weight smaller than the average?". See Listing 3 and 4.

\begin{listing}[h]%
\caption{Generated SQL Query}%
\label{lst:our_approach1}%
\begin{lstlisting}[language=SQL]
SELECT DISTINCT model_list.model
FROM model_list
JOIN car_names ON model_list.modelid = car_names.model
JOIN cars_data ON car_names.makeid = cars_data.id
WHERE cars_data.weight <
    (SELECT AVG(weight)
     FROM cars_data)
\end{lstlisting}
\end{listing}
\begin{listing}[h]%
\caption{Ground Truth (Gold) SQL Query}%
\label{lst:ground_truth1}%
\begin{lstlisting}[language=SQL]
SELECT T1.model
FROM CAR_NAMES AS T1
JOIN CARS_DATA AS T2 ON T1.MakeId = T2.Id
WHERE T2.Weight <
    (SELECT avg(Weight)
     FROM CARS_DATA)
\end{lstlisting}
\end{listing}

The generated SQL joins on the \verb|model_list| table, however the ground truth SQL (correctly) doesn't because the column \verb|model| is already in the \verb|car_names| table. Even when the database schema is given (Format shown in Figure \ref{fig2}) the LLM still geenrates the extra JOIN. This causes the generated SQL to return nothing as a output result table, while the ground truth returns 230 rows of data in its result table.

\subsection{\textit{Spider} Inconsistencies}
A key finding of our study is the demonstration of inconsistencies within the \textit{spider} dataset. During our experiments, we observed that a significant number of SQL queries generated through our LLM-based approach, which are semantically correct and yield result tables identical to those of the ground truth, are still being classified as incorrect. This includes some other instances where the LLM-generated queries are an exact match, yet are flagged as incorrect by the dataset's evaluation script. This discovery shows a critical flaw in the Spider's evaluation methodology and highlights the need for a more nuanced approach to assessing the accuracy of LLM-generated SQL queries.

When presented with the natural language question, "Show the stadium name and capacity with the most number of concerts in 2014 or later", both the generated SQL and the ground truth SQL produce identical results. The only distinction lies in the usage of table aliases; the ground truth employs aliases \verb|T1| and \verb|T2| for table names, whereas the generated SQL directly utilizes the actual names of the tables.  Many of the incorrect queries are exactly like this, making it extremely hard to tell when a query is actually incorrect or not.

\subsubsection{Incorrect Ground Truth}
In the \textit{spider} dataset, in a few instances that we could see, the generated SQL is more accurate than the dataset's provided ground truth. A notable example is found in entry 944 of the development set. The corresponding natural language question is: "What are the first name and last name of the professionals who have done treatment with cost below average?". The issue with the ground truth SQL lies in its use of a JOIN operation without specifying the joining condition. This oversight leads to an excessively broad result set. It is essential to join the \verb|Professionals| and \verb|Treatments| tables on the \verb|professional_id| field to accurately reflect the relationships in the data. Failing to specify this join condition results in the inclusion of professionals who have never received treatment, assigning them a default treatment cost of zero. Consequently, when comparing treatment costs against the average, these professionals erroneously appear in the results. This misalignment with the natural language query's intent indicates a critical flaw in the ground truth SQL provided in the dataset.

\subsection{Query Structure}
A significant portion of the SQL queries incorrectly generated by the LLMs falls into a category characterized by their unnecessary complexity and deviation from the simplicity of the ground truth solutions. Notably, in numerous instances, the LLMs resorted to using nested sub-queries to address problems that the ground truth solutions efficiently solved through straightforward table joins, and vice-versa. This complexity presents a substantial challenge in repairing these SQL queries. Due to their fundamentally incorrect structure, these queries are particularly resistant to correction through both LLM few-shot learning and algorithmic repair methods. In an analysis of forty randomly selected incorrect queries, nine exhibited this type of over complication, indicating it as one of the more prevalent and challenging issues in SQL generation via LLMs.

\section{Conclusions}
Testing and using these different approaches has provided clear insights to the process of using LLMs for Text-to-SQL program synthesis. It is now very clear that the closed-source models are still the leaders in high performing LLMs, with their ability to perform few-shot in-context learning \cite{brown2020fewshot}. This could be due to lower parameter counts in WizardLM's \textit{WizardCoder} model or just the underlying data it is created with. It should be noted that \textit{gpt-3.5-turbo} was just revealed to be only a 20B parameter model, not too much larger than \textit{WizardCoder}.

This research has yielded critical insights into the categorization of inaccuracies in SQL queries generated by LLMs. Despite their proficiency, LLMs encounter considerable challenges, particularly with the more complex entries in the \textit{spider} dataset. The errors predominantly fall into seven categories: incorrect selection or ordering of columns, erroneous grouping, inaccurate conditional value predictions, divergence in aggregate functions compared to the ground truth, inappropriate or insufficient JOIN clauses, discrepancies within the \textit{spider} dataset itself, and fundamentally incorrect query structures. Considering that a significant number of incorrect queries have a fundamentally flawed query structure, any post-processing step that aims to ``fix'' generated queries has to go beyond superficial features and consider the semantics of the generated query.

A notable observation is that many SQL queries, incorrectly flagged as erroneous, actually produce result tables identical to those generated by the ground truth SQL. In some cases, the ground truth SQL itself contains inaccuracies. These findings highlight the need for nuance in evaluating the accuracy of LLM-generated SQL queries.

\bigskip

\bibliography{bib}

\end{document}